# NoiseOut: A Simple Way to Prune Neural Networks

**Mohammad Babaeizadeh, Paris Smaragdis & Roy H. Campbell**
Department of Computer Science
University of Illinois at Urbana-Champaign
{mb2,paris,rhc}@illinois.edu.edu

## Abstract

Neural networks are usually over-parameterized with significant redundancy in the number of required neurons. This results into unnecessary computation and memory usage at inference time. One common approach to address this issue is to prune these big networks by removing extra neurons and parameters while maintaining the accuracy. In this paper, we propose NoiseOut, a fully automated pruning algorithm based on the correlation between activations of neurons in the hidden layers. We show that adding additional output neurons with fully random targets results into higher correlation between neurons which makes pruning by NoiseOut even more effective. Finally, we test our method on various networks and datasets. These experiments exhibit high pruning rates while maintaining the accuracy of the original network.

## 1 Introduction

Neural networks and deep learning recently achieved state-of-the-art solutions to many problems in computer vision [1, 2], speech recognition [3], natural language processing [4] and reinforcement learning [5]. Quite often, we see in such tasks the use of very large networks. Such oversized networks can easily overfit on the training dataset while having poor generalization on the testing data [6]. A rule of thumb for obtaining good generalization is to use the smallest number of parameters that can fit the training data [7]. Unfortunately, this optimal size is not usually obvious and therefore the size of the neural networks is determined by a few rules-of-thumb [8] which do not guarantee an optimal size. One common approach to overcome overfitting is to choose an over-sized network and then apply regularization [9] and Dropout [10]. However, these approaches do not reduce the number of parameters and therefore do not resolve the high demand of resources at test time.

Another method is to start with an oversized network and then use pruning algorithms to remove redundant parameters while maintaining the network's accuracy [11]. These methods need to estimate the upper-bound size of a network, a task for which there are adequate estimation methods [12]. If the size of a neural network is bigger than what is necessary, in theory it should be possible to remove some of the extra neurons without affecting its accuracy. In order to do so, one should find neurons which once removed result in no additional error. However, this may not be as easy as it sounds as all the neurons are contributing to the final prediction and removing them usually leads to error.

Our goal in this paper is two fold. First, we introduce a pruning method called NoiseOut based on the correlation between activations of the neurons. Since the effectiveness of this method hinges on high correlations between neuron outputs, we propose an approach which modifies the training cost function, and show that this modification increases the correlation of neuronal activations and thus facilitates more aggressive pruning.

## 2 Proposed Method

In this section, we describe the details of the proposed method called NoiseOut. First, we show how this method can prune a network and then how the pruning can be improved.

**Algorithm 1** NoiseOut for pruning hidden layers in neural networks

**procedure** TRAIN($X, Y$) ▷ X is input, Y is expected output
    $W \leftarrow initialize_weights()$
    **for each** iteration **do**
        $Y_N \leftarrow generate_random_noise()$ ▷ generate random expected values
        $Y' \leftarrow concatenate(Y, Y_N)$
        $W \leftarrow back_prop(X, Y')$
        **while** $cost(W) \leq threshold$ **do**
            $A, B \leftarrow find_most_correlated_neurons(W, X)$
            $\alpha, \beta \leftarrow estimate_parameters(W, X, A, B)$
            $W' \leftarrow remove_neuron(W, A)$
            $W' \leftarrow adjust_weights(W', B, \alpha, \beta)$
            $W \leftarrow W'$
    **return** $W$

### 2.1 Pruning a single neuron

In NoiseOut, instead of removing a neuron, we merge two neurons with highly correlated activations into one. The main rationale behind merging the most correlated pair of neurons is to keep the signals inside the network as close to the original network as possible. In an ideal scenario, the correlation between these two neurons is 1. Note that this assumption is an ideal case in which, removing one of the neurons results into no change in accuracy since the final output of the network will stay exactly the same. In non-ideal cases, when the highest correlated neurons are not fully correlated, merging them into one neuron may alter the accuracy. However, continuing the training after the removal may compensate for this loss. If this doesn't happen, it means the removed neuron was necessary to achieve the target accuracy and the algorithm cannot compress the network any further without accuracy loss.

NoiseOut simply repeats this process to compress the network. The pruning ends when the accuracy of the network drops below some given threshold. Note that the pruning process is happening while training. Algorithm 1 shows the final NoiseOut algorithm. For the sake of readability, this algorithm has been shown for networks with only one hidden layer. But the same algorithm can be applied to networks with more that one hidden layer by performing the same pruning on all the hidden layers independently. It can also be applied to convolutional neural networks that use dense layers, in which we often see over 90% of the network parameters [13].

### 2.2 Encouraging correlation between neurons

The key element for successful pruning of neural networks using NoiseOut is high correlation between activation of the neurons. Higher correlation between these activations means more effective pruning. However, there is no guarantee that back-propagation results in correlated activations in a hidden layer. In this section, we propose an adjustment to the cost function and show how it encourages increased correlation between redundant neurons.

Instead of changing the cost function directly, we add additional output nodes, called *noise outputs*. The targets for noise outputs will randomly change in each iteration based on a predefined random distribution. For simple networks it is possible to mathematically demonstrate the effect of adding noise outputs and how it intensifies the correlation between activation in the hidden layers which will subsequently make the pruning task more effective. We emit these proofs due lack of space in this short paper.

To demonstrate the effect of adding noise outputs, we experimented with adding noise outputs with different random distributions on a 2 layer MLP (2-2-1) and a 6 layer MLP (2-2-2-2-2-2-1). These noise distributions include *Gaussian*, *Binomial* and *Constant* (in this case the target of the noise outputs is a constant value). We also compared the results with the case when there is no noise output. As it can be seen in Figure 1, adding noise outputs helped the neurons to achieve higher correlation compared to a network with no noise output. Binomial noise acts chaotic at the beginning due its sudden change of expected values in the noise outputs while Gaussian noise improved the correlation the best in these experiments.

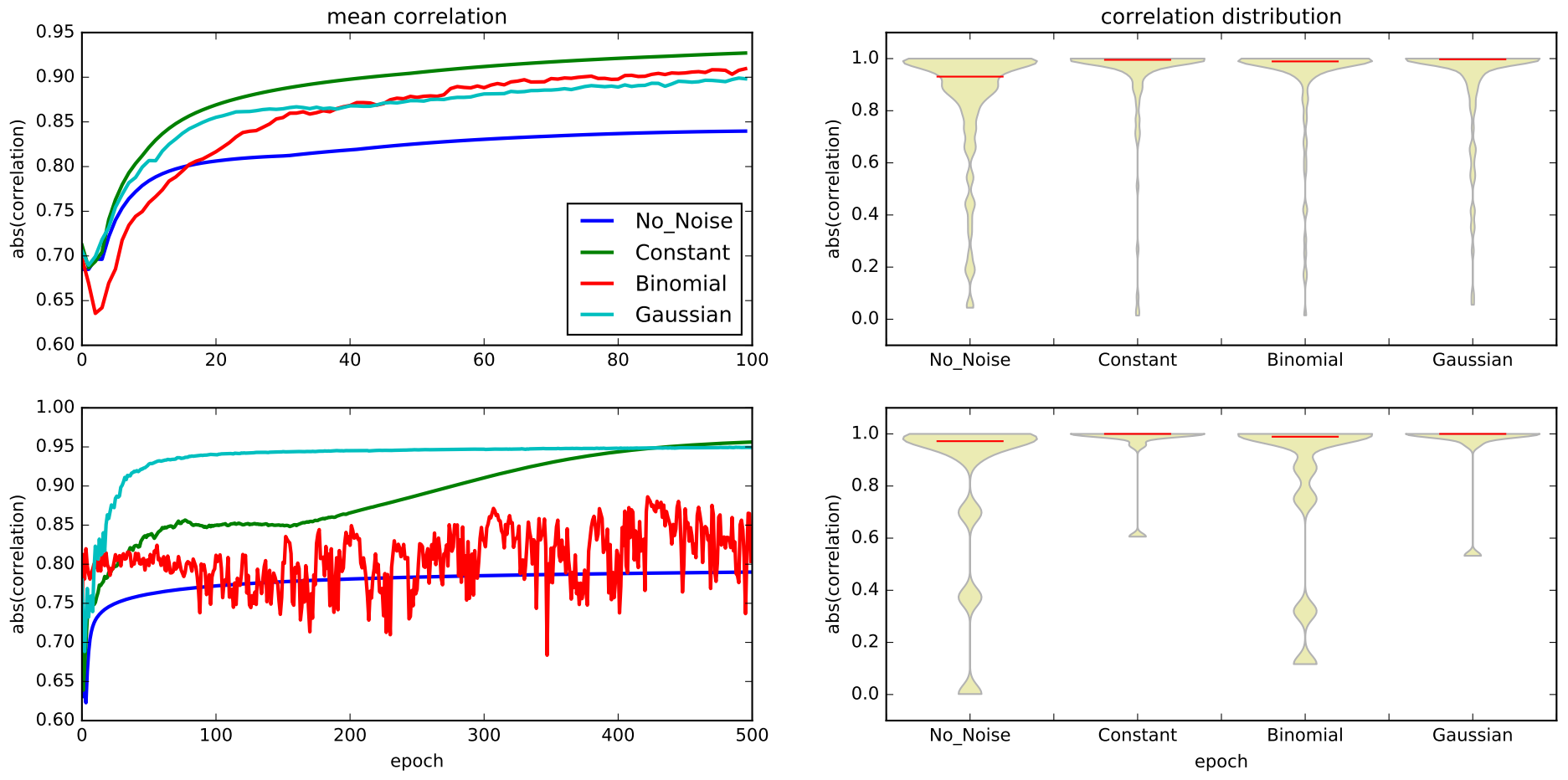

Figure 1: The effect of adding noise outputs to the correlation between the activation of neurons in the hidden layers. The top row shows the correlation of two hidden neurons in a 2 layer MLP while the bottom row is the correlation between the two neurons on the first hidden layer of a 6 layer MLP. In both cases, the left graph represents the mean correlation (in 100 runs) of these neurons during training and right graph is the distribution of these correlations. As it can be seen in these graphs, adding noise outputs improves the correlation between neurons in the hidden layer. Note that in all of these graphs, the absolute value of correlation has been presented.

# 3 Experiments

We implemented NoiseOut on Keras [14] and pruned different networks on different dataset: Lenet-300-100 and Lenet-5 on MNIST [15] image dataset and a convolutional neural network with one dense layer on SVHN [16]. For each one of these networks, we ran experiments with different random distributions as $P_{Y_N}$.

## 3.1 MNIST

### 3.1.1 Lenet-300-100

Lenet-300-100 is a fully connected network with two hidden layers, with 300 and 100 neurons each, which achieves 3.05% error rate on MNIST [17]. We pruned this network using NoiseOut with Gaussian distribution and multiple accuracy thresholds. The result of these experiments is shown in Figure 2. As it is evident, lower accuracy thresholds result into more pruned parameters while the gap between training and testing threshold stays the same. This shows that pruning the network using NoiseOut does not lead to overfitting.

In order to examine the effect of different random distributions on NoiseOut efficiency, we also tested this algorithm with multiple distributions and the same accuracy threshold. The result of this experiments is demonstrated at Table 1.

### 3.1.2 LeNet-5

LeNet-5 is a convolutional network with two convolutional layers and one hidden fully connected layers which achieves 0.95% error rate on MNIST [17]. Over 98% of the total number of parameters are in the dense layer and pruning them can decrease the model size significantly. We repeated the same experiments as what has been described for LeNet-300-100. The results of these experiments can be seen in Figure 2 and Table 3. NoiseOut with Gaussian noise manages to remove to 97.75% of the weights and achieved error rate of 0.95% with only 3 neurons in the hidden layer. This reduces the total number of weights in LeNet-5 by a factor of 44.

## 3.2 SVHN

For SVHN data set, we used a deep convolutional neural network with over 1 million parameters which achieves 93.39% and 93.84% accuracy on training set and test set respectively.

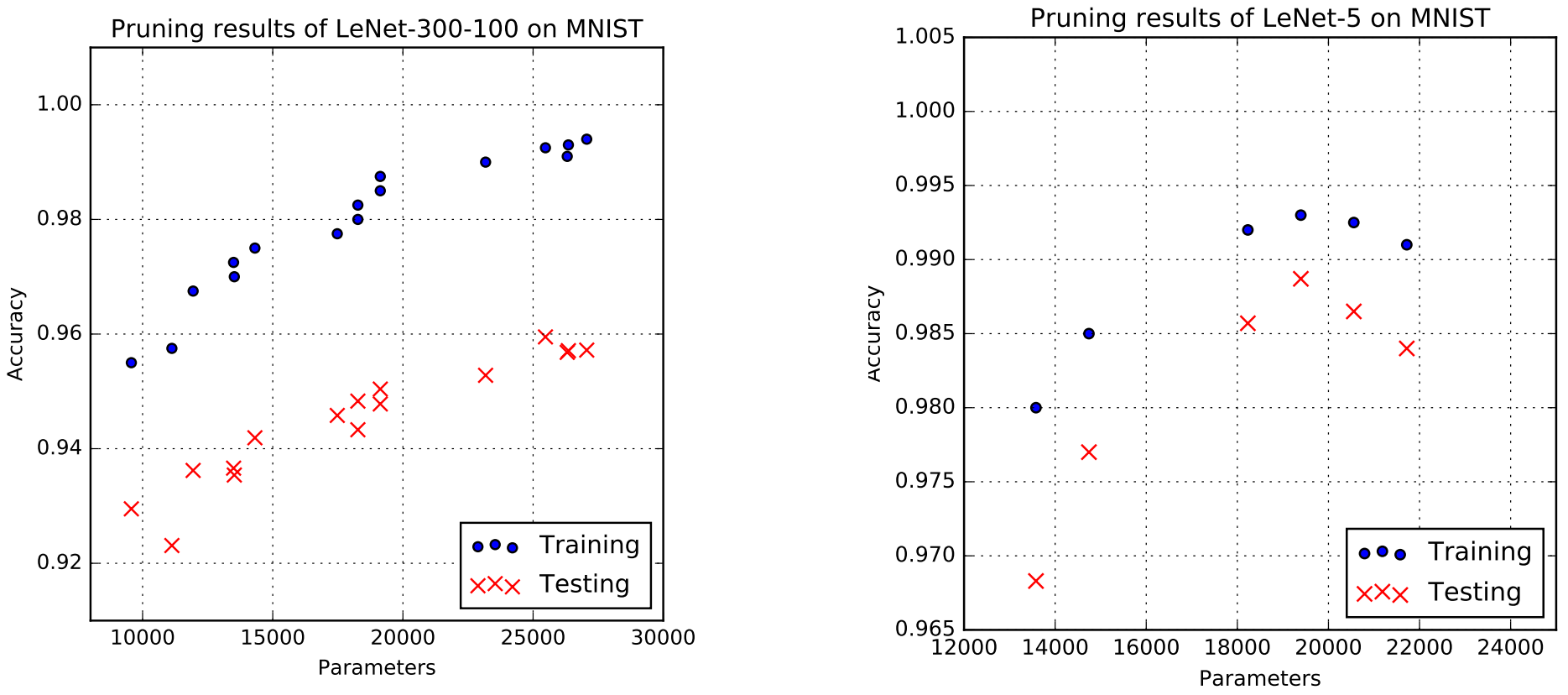

Figure 2: Pruning LeNet-300-100 and LeNet-5 on MNIST data set with various accuracy thresholds. $x$ axis represents the total number of parameters in the pruned network (including weights in the convolutional layers), while $y$ axis shows the accuracy of the model on test and training dataset.

Table 1: Pruning Lenet-300-100 on MNIST. In all of the experiments the error rate is 3.05%

| Method | Noise Neurons | Layer 1 Neurons | Layer 2 Neurons | Parameters | Removed Parameters | Compression Rate |
|---|---|---|---|---|---|---|
| Ground Truth | - | 300 | 100 | 266610 | - | - |
| No_Noise | - | 23 | 14 | 15989 | 94.00% | 16.67 |
| Gaussian | 512 | 20 | 9 | 15927 | 94.02% | 16.73 |
| Constant | 512 | 20 | 7 | 15105 | 94.33% | 17.65 |
| Binomial | 512 | 19 | 6 | 11225 | 95.78% | 23.75 |
| No_Noise | - | 13 | 12 | 10503 | 96.06% | 20.89 |
| Gaussian | 1024 | 16 | 7 | 12759 | 95.21% | 18.58 |
| Constant | 1024 | 18 | 7 | 14343 | 94.62% | 17.61 |
| Binomial | 1024 | 19 | 7 | 15135 | 94.32% | 25.38 |

In this experiment, we set the cut accuracy to the same accuracy as base network. As it can be seen in Table 2, NoiseOut pruned more than 85% of the parameters from the base model while maintaining the accuracy.

## 4 Conclusion

In this paper, we have presented NoiseOut, a simple but effective pruning method to reduce the number of parameters in the dense layers of neural networks by removing neurons with correlated activation during training. We showed how adding noise outputs to the network can increase the correlation between neurons in the hidden layer and hence result to more effective pruning. The experimental results on different networks and various datasets validate this approach, achieving state-of-the-art compression rates without loss of accuracy.

Table 2: Pruning a convolutional network trained on SVHN dataset with 93.39% accuracy

| Method | Dense Layer Neurons | Parameters | Removed Parameters |
|---|---|---|---|
| Ground Truth | 1024 | 1236250 | - |
| No_Noise | 132 | 313030 | 74.67% |
| Gaussian | 4 | 180550 | 85.39% |
| Constant | 25 | 202285 | 83.63% |
| Bionomial | 17 | 194005 | 84.30% |

Table 3: Pruning Lenet-5 on MNIST. In all of the experiments the error rate is 0.95%

| Dense Layer Neurons | Parameters | Removed Parameters |
|---|---|---|
| 512 | 605546 | - |
| 313 | 374109 | 38.21% |
| 3 | 13579 | 97.75% |
| 33 | 48469 | 91.99% |
| 26 | 40328 | 93.34% |